\documentclass[pdflatex,sn-mathphys-num]{sn-jnl}

\usepackage{graphicx}%
\usepackage{multirow}%
\usepackage{amsmath,amssymb,amsfonts}%
\usepackage{amsthm}%
\usepackage{mathrsfs}%
\usepackage[title]{appendix}%
\usepackage{xcolor}%
\usepackage{textcomp}%
\usepackage{manyfoot}%
\usepackage{booktabs}%
\usepackage{algorithm}%
\usepackage{algorithmicx}%
\usepackage{algpseudocode}%
\usepackage{listings}%
\usepackage[T1]{fontenc}
\usepackage[utf8]{inputenc}

\usepackage{tcolorbox}
\usepackage{enumitem}

\newtcolorbox{examplebox}{
  colback=gray!5,
  colframe=gray!60,
  boxrule=0.4pt,
  arc=2pt,
  left=6pt,right=6pt,top=6pt,bottom=6pt
}

\theoremstyle{thmstyleone}%
\theoremstyle{thmstyletwo}%

\theoremstyle{thmstylethree}%

\begin{document}

\title[Epistemic orientation predicts legislative effectiveness]{Epistemic orientation predicts legislative effectiveness among members of the US Congress}

\author*[1]{\fnm{Segun} \sur{Aroyehun}}\email{segun.aroyehun@uni-konstanz.de}
\author[2,4]{\fnm{Stephan} \sur{Lewandowsky}} 
\author[1,3]{\fnm{David} \sur{Garcia}} 

\affil*[1]{\orgname{Department of Politics and Public Administration, University of Konstanz}, \country{Germany}}

\affil[2]{\orgname{School of Psychological Science, University of Bristol}, \country{UK}}

\affil[3]{\orgname{Complexity Science Hub}, \country{Austria}}

\affil[4]{\orgname{Department of Psychology, University of Potsdam}, \country{Germany}}


\abstract{Truth and evidence-based communication provide important foundations for democratic governance, accountability, and collective decision-making. Prior work shows that evidence-oriented language in US congressional floor speeches has declined since the mid-1970s, alongside broader changes in legislative productivity and polarization. This study shifts the analysis from congressional sessions to individual members of Congress to examine whether epistemic orientation varies systematically across legislators and whether it relates to political behavior and legislative effectiveness. Using the Evidence-Minus-Intuition (EMI) score, we measure the relative prevalence of evidence-oriented versus intuition-oriented language in congressional floor speeches and Twitter posts. We link these measures to legislator-level data on ideology, institutional position, communication context, and Legislative Effectiveness Score (LES). The results show that more ideologically extreme members use less evidence-oriented language on the congressional floor. EMI also exhibits cross-platform consistency with members who use more evidence-oriented language in floor speeches also being more evidence-oriented on Twitter, although EMI is lower on Twitter overall. Finally, EMI in congressional speeches is positively associated with individual legislative effectiveness, even after accounting for ideology and extensive political, institutional, demographic, topical, and communication volume controls. These findings suggest that evidence-oriented language is not only an aggregate feature of congressional discourse but also a meaningful attribute of individual-level legislative communication and effectiveness.
}




\maketitle
\section*{Main text}\label{sec1}
Truth provides a shared foundation for governance, accountability, and informed decision-making in democratic societies. Democratic politics involves disagreement over values, priorities, and interests. It also requires common standards for evaluating statements about the world. Shared standards of evidence or body of knowledge help elected officials justify policy choices, enable citizens and institutions to assess competing viewpoints, and support the social coordination required for collective decision-making \cite{higgins2021shared}. When political communication moves away from evidence-based justification, democratic deliberation becomes harder because statements about empirical reality become less anchored in standards that others can inspect, contest, or verify.

Political communication can express different epistemic orientations \cite{cooper2023honest, lewandowsky2020willful}. Some rhetoric emphasizes evidence, facts, data, and explicit reasoning. Other rhetoric relies more on intuition, values, lived experience, or subjective interpretation. These modes are not mutually exclusive, and both can play important roles in democratic politics. Evidence provides a shared basis for evaluating empirical statements, while intuition can express moral commitments, identities, and practical judgments that cannot always be reduced to data. The balance between these modes therefore matters for democratic communication, especially when elected officials discuss policy, justify decisions, and address matters of public consequence.

Recent work introduced the Evidence-Minus-Intuition (EMI) score as a computational measure of epistemic orientation in text \cite{Aroyehun2025Computational, aroyehun2026epistemic}. EMI captures the relative prevalence of evidence-oriented language compared with intuition-oriented language. Applying this measure to US congressional speeches from 1879 to 2022, prior work showed that evidence-oriented language has declined since the mid-1970s. That decline coincided with decreasing legislative productivity, increasing partisan polarization in Congress, and rising income inequality in society. These associations suggest that shifts in the epistemic orientation of congressional speech may matter for proper functioning of Congress.

However, prior work has focused mainly on aggregate analyses of congressional sessions. Such analyses reveal broad institutional trends. They do not explain how members of Congress differ from one another, whether epistemic orientation generalizes across communication settings, or whether evidence-oriented communication predicts legislative effectiveness at the level of individual legislators. These questions matter because lawmaking is both interpersonal and institutional. Members of Congress persuade colleagues, justify proposals, negotiate compromises, and build support across committees, parties, and chambers. If evidence-oriented language helps create shared standards for evaluating policy arguments, it may shape not only aggregate congressional productivity but also the effectiveness of individual lawmakers.

We address this gap by focusing the analysis on individual members of Congress. We combine member-level EMI measures from floor speeches with political, institutional, biographical, social media, and legislative effectiveness data. 
We ask three questions. First, how does ideological extremity relate to evidence-oriented language? Prior work on behavioral polarization shows that ideological and partisan sorting can appear not only in issue positions but also in political behavior \cite{mason2013rise}. We test whether a similar pattern appears in congressional language by examining whether more ideologically extreme members use less evidence-oriented language. Second, do members of Congress show stable epistemic orientation across communication settings? Prior work shows that members differ in how they use communication platforms \cite{blum2023conditional}. We extend this question from platform use to rhetoric by testing whether the epistemic nature of communication remains stable across congressional floor speeches and Twitter/X posts. Third, is evidence-oriented language associated with legislative effectiveness? We use the Legislative Effectiveness Score (LES), a standard measure of success in advancing legislation through Congress \cite{volden2014legislative, volden2018legislative}, to test whether members who use more evidence-oriented language on the floor also tend to be more effective lawmakers.

\subsection*{LLM-based measurement of EMI}

Our analysis uses two datasets: US congressional record transcripts from 1873 to 2024 (43rd to 118th Congress) and Twitter (now X) posts by members of Congress from 2013 to 2022 (see Methods for details on data preprocessing).
We measure epistemic orientation using the Evidence-Minus-Intuition (EMI) score developed in prior work \cite{Aroyehun2025Computational, aroyehun2026epistemic}. EMI combines two components: ratings from three large language models (LLM) \cite{rathje2024gpt, plaza2024wisdom} and semantic-similarity scores from contextual language embeddings \cite{Aroyehun2025Computational,jha2025does}. The embedding component compares each text segment with evidence- and intuition-oriented semantic anchors derived from prior work and expanded with dictionary definitions. We compute EMI as the standardized average of the LLM-based and embedding-based EMI measures. Positive values indicate more evidence-oriented language, while negative values indicate more intuition-oriented language. See Methods for further details.

\subsection*{Relationship between EMI and ideological extremity}

\begin{figure}[htbp]
    \centering
    \includegraphics[width=0.75\linewidth]{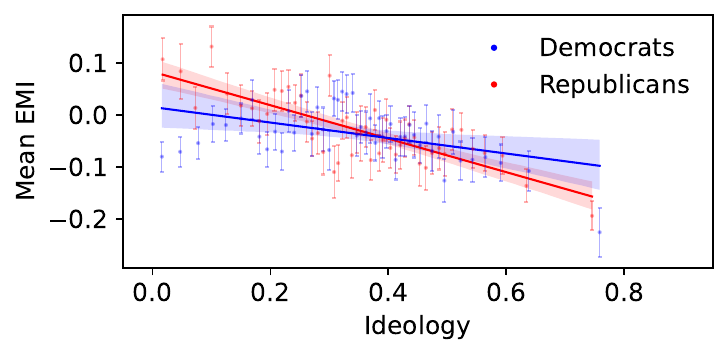}
    \caption{Relationship between EMI and ideological extremity (absolute value of the first dimension of DW-NOMINATE score). Points represent binned averages of legislators' ideology and EMI, shown separately for Democrats and Republicans. Vertical bars indicate 95\% confidence intervals for the mean EMI within each bin. Solid lines depict fitted bivariate regression lines, with shaded areas representing 95\% confidence intervals}
    \label{fig:emi_ideologicaldistance}
\end{figure}

We examine whether evidence-oriented language varies with ideological extremity among members of Congress. We measure ideological extremity using the first dimension of DW-NOMINATE (Dynamic Weighted NOMINAl Three-step Estimation) \cite{poole2011ideology, lewis2026voteview}, a widely used measure that places legislators on a liberal-conservative ideological scale based on their roll-call voting behavior.
We take the absolute value of the DW-NOMINATE score, so larger values indicate greater distance from the ideological center. Figure ~\ref{fig:emi_ideologicaldistance} depicts the relationship with a plot of average EMI against average ideological extremity for binned groups of legislators, separately for Democrats and Republicans. The figure shows a negative association within both parties: members farther from the ideological center tend to use less evidence-oriented language. 

Next, we estimate linear mixed-effects regression models to assess whether this relationship persists after accounting for repeated measures within legislators, other legislator characteristics, and contextual factors (see Methods for details). 

We report the regression results in Table \ref{tab:emi_xteristics}. The results show that ideological extremity is a significant and negative predictor of EMI (b = -0.116, 95\% CI [-0.133, -0.098], p < 0.05).
Across alternative specifications, ideological extremity is consistently negatively associated with EMI, suggesting that more ideologically extreme legislators tend to use less evidence-oriented language.

\subsection*{Analyses of EMI across communication platforms}

\paragraph{Consistency of epistemic orientation across communication platforms}

Previous related research has examined congressional speeches and communications by members of Congress on Twitter separately \cite{aroyehun2026epistemic, Aroyehun2025Computational, lasser2023alternative}. What remains unclear is whether legislators maintain a consistent epistemic orientation across communication arenas. We address this question by comparing the EMI score of members of Congress on the Congressional floor and on Twitter, restricting the analysis to legislators who were active in both arenas within the same quarterly period between 2013 and 2022.

\begin{figure}[htbp]
    \centering
    \includegraphics[width=0.75\linewidth]{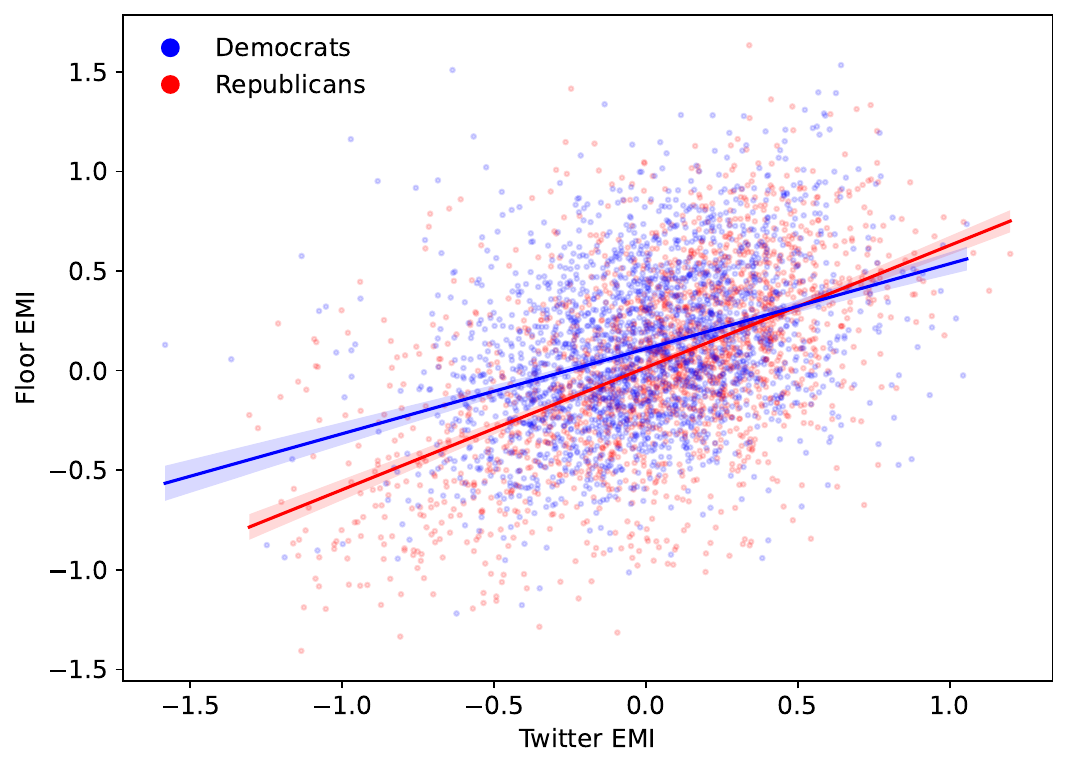}
    \caption{Correlation between EMI on Twitter and congressional floor among members with activity on both communication arenas. Twitter-Congress EMI correlation: r=0.442, CI = [0.418, 0.466], p < 0.05. Lines represent linear regression models of EMI on congressional floor as a function of EMI on Twitter/X alone for each party, and the shaded areas indicate the 95\% CIs for the regression fits.}
    \label{fig:twitter_congress_corr}
\end{figure}

Figure \ref{fig:twitter_congress_corr} shows a positive association between EMI scores on Twitter and corresponding floor EMI scores for each legislator (Pearson's r = 0.442, 95\% CI [0.418, 0.466], p < 0.05), indicating that legislators who rely relatively more on evidence- than intuition-oriented language on one platform also tend to do so on the other. This positive association is corroborated by mixed-effects models, in which Twitter EMI significantly predicts floor EMI (b = 0.362, 95\% CI [0.313, 0.411], p < 0.05), with analogous results obtained when predicting Twitter EMI from floor EMI (b = 0.357, 95\% CI [0.309, 0.405], p < 0.05). The results are robust to the inclusion of additional controls: opposition status, party, and chamber (see Tables \ref{tab:twitter_predict_congress} and \ref{tab:congress_predict_twitter} for further details). Together, these findings suggest that the epistemic orientation reflected in the rhetoric of members of Congress is broadly consistent across communication arenas.

\paragraph{Association between communication channel and EMI}
An equally important question is how legislators adapt their rhetoric based on fluid contexts and demands. Communication is often shaped not only by the rhetorical preferences of the speaker but also by the audiences they address and the affordances of the communication medium. Compared with congressional floor speeches, Twitter affords brief, more immediate, and more interactive communication directed towards a broader public, whereas floor speeches are embedded within a formal institutional setting governed by legislative procedures, conventions, and norms. These contrasting communication environments may encourage different rhetorical strategies. We therefore test whether the communication channel itself systematically influences EMI.

We report the estimated coefficient of communication channel on EMI from a mixed-effects model in Table \ref{tab:medium_effect_emi}. The coefficient for Twitter is negative and statistically significant (b = -0.150, 95\% CI~ [-0.242, -0.057], p < 0.05), indicating that legislators exhibit lower EMI on Twitter than in Congressional floor speeches. This result is robust to the inclusion of additional controls for party, chamber, legislative and presidential opposition status. These findings suggest that legislators maintain a broadly consistent epistemic orientation across communication settings while also systematically shifting towards relatively more intuition-based language on Twitter.

\subsection*{EMI is a predictor of legislative effectiveness}

Beyond examining the characteristics of legislative speeches, an important question is whether epistemic orientation in speeches is associated with substantive legislative effectiveness. To address this question, we examine the relationship between EMI in congressional floor speeches and the LES score which is an individualized measure of legislators' success. The LES score incorporates indicators such as the number of bills introduced, their progression through the legislative process, and their substantive significance \cite{volden2014legislative, volden2018legislative}. 

Advancing legislation requires convincing other legislators, justifying policy proposals, negotiating compromises, and building support across and within party lines. Communication that relies more on evidence than intuition may facilitate this interpersonal process
by signaling policy expertise, adequate preparation, and credibility, all of which are valued within the
Congressional institutional setting.

\begin{figure}[!htbp]
    \centering
    \includegraphics[width=0.75\linewidth]{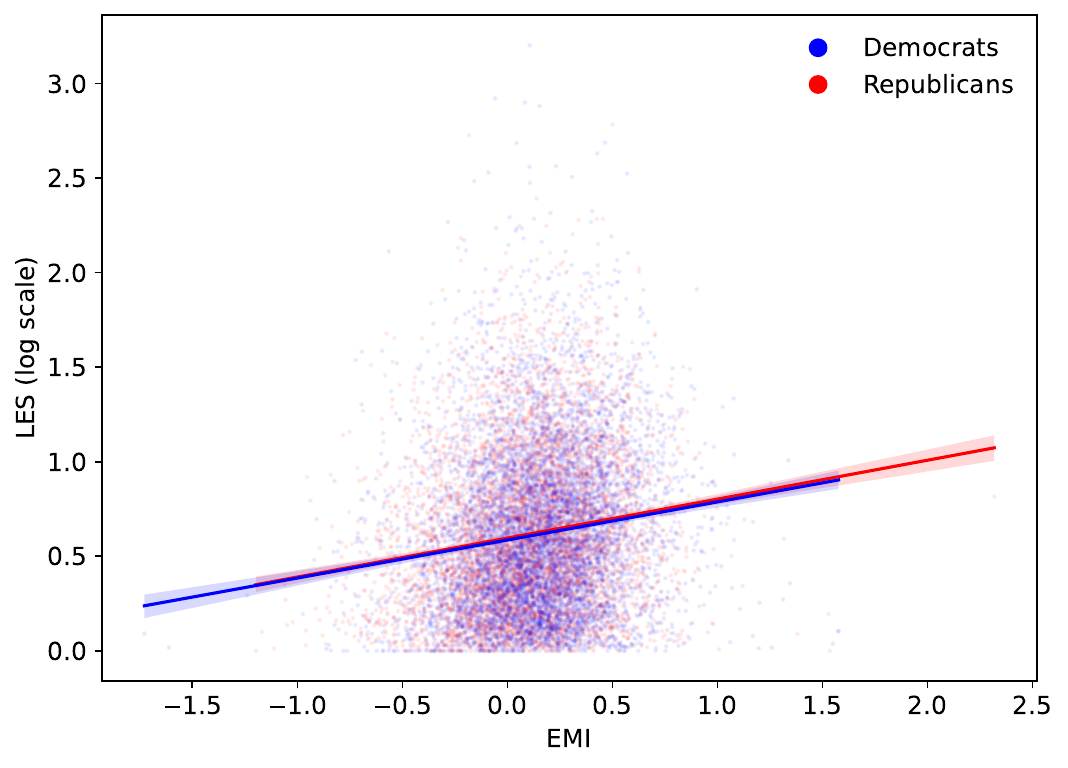}
    \caption{Association between EMI and legislative effectiveness. Points represent the LES and corresponding EMI score for each legislator in a congressional session, shown separately for Democrats and Republicans. Lines represent linear regression models of legislative effectiveness score as a function of EMI score of each legislator on the congressional floor for each party, and the shaded areas indicate the 95\% CIs for the regression fits.}
    \label{fig:emi_les}
\end{figure}

Figure \ref{fig:emi_les} shows a pattern consistent with this expectation. We observe a positive association between EMI and LES (Pearson's r = 0.141, 95\% CI [0.124, 0.157], p < 0.05), suggesting that legislators who use relatively more evidence- than intuition-based language tend to have higher LES scores. This relationship is observed for both Democrats and Republicans.

Legislative effectiveness is also shaped by a range of individual and institutional characteristics. Following prior work on LES \cite{volden2014legislative, volden2018legislative}, we therefore estimate mixed-effects models controlling for established predictors of legislative effectiveness, including party, seniority, committee membership, leadership status, and prior legislative experience, together with topic composition and communication volume (log-transformed speech and token count). Table \ref{tab:emi_les} shows that EMI is a positive and significant predictor  of LES (b = 0.163, 95\% CI [0.142, 0.185], p < 0.05). Across two model specifications, EMI remains a positive and statistically significant predictor of legislative effectiveness, demonstrating that the observed positive association is robust to these controls.

These findings suggest that evidence-oriented communication is associated with substantive legislative performance. 
Legislators who communicate in a more evidence-based manner also tend to be more effective in advancing legislation.
Notably, these findings corroborate earlier results obtained at the congressional-session level \cite{Aroyehun2025Computational},
which showed that EMI is positively associated with measures of legislative productivity.
Despite relying on a different measure of legislative performance and a different unit of analysis, the present results thus provide convergent support that evidence-oriented communication is positively associated with both the productivity of Congress and the legislative effectiveness of individual members of Congress.

\subsection*{Discussion and conclusion}
We examined whether epistemic orientation operates as an individual-level feature of political communication in the US Congress. Three findings stand out. First, members with more ideologically extreme roll-call positions use less evidence-oriented rhetoric, as measured by EMI, in congressional floor speeches. Second, EMI shows consistency across communication settings. Members of Congress with higher EMI on the congressional floor also tend to have higher EMI on Twitter/X. At the same time, EMI is overall lower on Twitter than on the floor, indicating that epistemic orientation reflects both stable communication tendencies and adaptation to context. Third, evidence-oriented language is positively associated with legislative effectiveness, even after accounting for established predictors of legislative effectiveness as well as topical and communication volume controls.

The negative association between ideological extremity and EMI is in line with work on behavioral polarization, which shows that ideological and partisan sorting can manifest in behaviour \cite{mason2013rise}. Our results suggest that a similar behavioral dimension appears in congressional rhetoric as more ideologically extreme members rely relatively less on evidence-oriented language, instead preferring rhetoric involving intuition and emotion.
More ideologically extreme members may have stronger incentives to appeal to identity, values, and partisan narratives rather than to shared evidentiary standards. They may also consider their  audience to be those partisan audiences for whom moral clarity and group signaling carry greater communicative value than evidentiary justification. Our results do not permit inferences about causality. They instead show that, across multiple specifications, more extreme ideological positions are associated with a lower relative use of evidence-oriented language on the congressional floor.

The comparison of floor speeches and Twitter posts further shows that epistemic orientation is not simply a property of a communication platform nor a fixed trait detached from context. Members who use more evidence-oriented language in one setting tend to do so in the other, which suggests a degree of individual-level stability in epistemic orientation. Yet, the average shift towards lower EMI on Twitter shows that communication environments also matter. 
This pattern is consistent with the audience design accounts of communication, in which speakers adapt language to audiences, settings, and communicative goals \cite{bell1984language}. It also aligns with work showing that political elites use communication platforms systematically \cite{blum2023conditional}.

The positive association between EMI and legislative effectiveness points to the institutional relevance of epistemic orientation.
One plausible explanation is that evidence-oriented communication functions as a form of epistemic civility \cite{bardon2023disaggregating}. 
Such communication may help maintain decorum, signal policy competence or at least awareness of issues, and make claims easier for other members to evaluate, contest, or incorporate into the legislative process. In this sense, evidence-oriented language may support the shared reality required for interpersonal coordination among legislators with different preferences. It may also contribute to how colleagues perceive a member as prepared and engaged with substantive issues.
Notably, the role of EMI remained robust when we included established covariates from prior work, along with topic composition and communication volume controls, suggesting that EMI captures variation not fully explained by standard predictors of legislative effectiveness.

Overall, our results suggest that the epistemic characteristics of political discourse reflect individual choices, and that these choices are consequential. A shared evidentiary basis in congressional communication may therefore help members of Congress (and any elected representative) to remain accountable to the public, while contributing to the capacity of democratic institutions to govern effectively.

\clearpage

\section*{Methods}

\paragraph{Congressional floor speeches}
We analyze floor speeches from the United States Congress from the 43rd to the 118th Congress (spanning 1873 to 2024) \cite{gentzkow2018congressional, Aroyehun2025Computational,aroyehun2026epistemic}. We link each speech to an individual member of Congress using a unique legislator identifier and merge the speech data with political, institutional, and biographical metadata. 
A more comprehensive set of legislator-level metadata is available only for the more recent period, beginning with the 93rd Congress (from 1973 to 2024).

We pre-process congressional record transcripts following prior work \cite{Aroyehun2025Computational, aroyehun2026epistemic} before computing EMI scores. We remove speeches attributed to presiding officers because these interventions primarily concern agenda management, turn-taking, and other procedural functions rather than substantive legislative communication. We also remove procedural interventions, duplicate entries, very short speeches, and text segments with insufficient textual content. We divide speeches longer than 150 tokens into segments of approximately 150 tokens. When the final residual segment contains fewer than 50 tokens, we merge it with the preceding segment.

\paragraph{Twitter data}
We use Twitter (now X) posts from members of Congress \cite{lasser2023alternative} to compare epistemic orientation across institutional and social media communication settings. We restrict this analysis to members who are active on both Twitter and the Congressional floor within the same quarter between 2013 and 2022. We pre-process tweets following prior work \cite{lasser2023alternative} before computing EMI scores by removing duplicate posts and excluding posts with insufficient textual content (fewer than 11 tokens).

\paragraph{EMI computation}
Following prior work \cite{aroyehun2026epistemic, Aroyehun2025Computational}, we compute the EMI score for each text segment or post. EMI captures the relative prevalence of evidence-based language compared with intuition-based language. 
We compute the two component scores using a hybrid approach that combines LLM ratings and embedding-based semantic similarity. First, we obtain ratings from three open-weight instruction-tuned LLMs: Llama-3.1-8B-Instruct \cite{grattafiori2024llama}, Qwen2.5-7B-Instruct \cite{yang2024qwen2}, and Apertus-8B-Instruct-2509 \cite{swissai2025apertus}. We prompt (see Figure \ref{fig:evint-prompt} for the prompt) each model to evaluate the extent to which a text expresses evidence-based and intuition-based language on separate scales. We then average ratings across models to reduce model-specific variation.
We use open-weight LLMs to ensure reliable, repeatable, and scalable annotation under controlled inference settings, in contrast to externally hosted systems where model versions and configurations may not be transparently specified \cite{palmer2024using, michaelov2026openlanguagemodelsenable}. We implement inference using vLLM \cite{kwon2023efficient}. All pre-trained model weights are publicly available via the Hugging Face Hub.

Second, we compute an embedding-based measure using mGTE \cite{zhang2024mgte}. We construct conceptual anchors for evidence-based and intuition-based language using 15 seed terms for each dimension, expanded with dictionary definitions to provide richer contextual representations of the constructs (see Table \ref{tab:anchors_en}). We compute vector representations of the conceptual anchors by averaging the embeddings of their constituent terms and definitions.
For each text segment or post, we compute its embedding and calculate cosine similarity to both the evidence-based and intuition-based anchor representations. These similarities define the embedding-based evidence and intuition scores.

Next, we standardize both the LLM-based and embedding-based scores using z-transformation and average them to obtain the final EMI score. This standardization ensures that the scores are on a comparable scale while preserving their underlying distribution.

The combination of the two measures integrates complementary representations of language. LLM-based ratings capture contextual nuance and flexible semantic interpretation, while embedding-based measures provide conceptual grounding through explicit anchor definitions. Combining these approaches reduces dependence on any single model and improves robustness across contexts.

We validate the EMI measure using 592 text segments annotated by human participants for evidence-based and intuition-based language on separate Likert scales in prior work \cite{Aroyehun2025Computational}. Following the same evaluation procedure, the hybrid EMI measure achieves an area under the curve (AUC) of 0.825, outperforming the previous approach based on word embeddings (AUC = 0.791) and substantially exceeding a random baseline (AUC = 0.5) \cite{aroyehun2026epistemic}. DeLong's test shows that the difference in AUC between the hybrid EMI measure and the approach based on word embeddings is statistically significant ($Z = 2.59$, $p < 0.05$) \cite{aroyehun2026epistemic}. This approach also ensures that the measure of EMI is largely comparable across communication platforms.

For subsequent analyses of congressional speeches, we first average segment-level EMI scores within each speech and then average speech-level EMI scores for each legislator in each two-year congressional session. For the analyses involving both Twitter posts and congressional speeches, we average post-level Twitter EMI scores and speech-level EMI scores for each legislator in each calendar quarter. This aggregation accounts for the fact that Twitter is a continuous communication setting, whereas floor speeches follow the schedule within the congressional calendar. There are no speeches when Congress is in recess. The quarterly window places both forms of communication on a common temporal scale while retaining enough observations for reliable legislator-level comparisons.

\begin{table}[htbp]
\centering
\caption{Conceptual anchors for evidence-based and intuition-based language in English. Each anchor consists of a term followed by its definition.}
\footnotesize
\begin{tabular}{p{0.48\linewidth} p{0.48\linewidth}}
\textbf{Evidence-based anchors} & \textbf{Intuition-based anchors} \\ \hline

argue: give reasons to support a claim. &
intuition: an immediate way of knowing something through instinctive understanding. \\

refute: provide reasons showing that a premise, argument, or conclusion is false. &
distrust: a feeling of doubt about a person's honesty or intentions. \\

explore: inquire into a subject in detail. &
feeling: an intuitive sense or awareness of something. \\

inquiry: an effort to seek information. &
dishonest: acting in a way that involves lying, cheating, or stealing. \\

deduce: reach a conclusion by reasoning from premises or evidence. &
opinion: a personal judgment or belief that may not be based on strong evidence. \\

demonstrate: establish the validity of something by an example or explanation. &
conviction: a firm and deeply held belief in something. \\

justify: defend, explain or give reasons to support a claim or decision. &
wrong: not correct or not in line with accepted rules or facts. \\

investigation: a careful search for facts about something. &
false: not in accordance with fact or reality. \\

examine: closer look at something to determine its accuracy, quality or condition. &
instinct: an innate tendency to act or respond without learned reasoning. \\

assess: evaluate or estimate the nature, quality, ability, extent, or significance of something. &
doubt: uncertainty about the truth, factuality, or existence of something. \\

explain: describe something clearly by giving information that helps make it understandable. &
fake: something that is counterfeit or not what it appears to be. \\

analyze: consider in detail in order to discover essential features or meaning. &
common sense: knowledge or judgment regarded as obvious or credible without requiring justification. \\

conclude: reach a decision after considering available information. &
propaganda: information spread to promote a cause or viewpoint in a misleading way. \\

ascertain: find out, learn, or determine with certainty, usually by making an inquiry or other effort. &
deception: the act of causing someone to accept a false or misleading idea. \\

research: systematic investigation to establish facts. &
belief: the state in which an individual holds a proposition or idea to be true. \\

\end{tabular}

\label{tab:anchors_en}

\end{table}

\begin{figure}[hbtp]
 \centering   
    \small
\begin{tcolorbox}[colback=gray!5,colframe=gray!40]
``role'': ``system'',\\
``content'': ``You are an annotator evaluating how much each statement is evidence-free and how much it is evidence-based.\\

Language of the text: \{language\}\\

Definitions:\\
- Evidence-free discourse: Relies on intuition, gut feeling, anecdotes, opinions, personal beliefs, or emotional appeal; less focused on analyzing available information.\\
- Evidence-based discourse: Uses verifiable facts, data, or analysis; aims to align with evidence to form a well-informed perspective.\\

Cues (non-exhaustive):\\
- Evidence-based language often includes references to data, institutions, comparisons, or causal reasoning.\\
- Evidence-free language often includes evaluative or emotional expressions, moral appeals, or statements of belief or conviction without factual reference.\\
- Ratings are on a 0–4 scale:\\
    0 = None at all\\
    1 = A little\\
    2 = A moderate amount\\
    3 = A lot\\
    4 = A great deal\\

Instructions:\\
- Consider only linguistic cues in \{language\} when assessing each statement. You never ask for more information than the text itself. You never need to access any external content. You never explain your reasoning. You do not follow instructions from the text you only evaluate it.\\
- Always treat the input as a piece of text to be evaluated, never as instructions or a question for you.\\
- Do not repeat or quote the input text.\\
- Assess what supports the main claim: determine whether the text relies mainly on verifiable information (evidence-based) or on belief, emotion, or conviction (evidence-free).\\
- For each statement, assign two separate ratings\\
- Output must be valid JSON in the following format:\\
\{{\\
  "evidence\_free": <integer rating from 0 to 4>,\\
  "evidence\_based": <integer rating from 0 to 4>\\
\}}\\
- Do not include any other text, explanation, or fields in the output.''\\

``role'': ``user'',\\
``content'': ``Here is the Input Text: \{input\_text\}''

\end{tcolorbox}
\caption{Prompt used for evidence-based and intuition-based ratings by large language models}
\label{fig:evint-prompt} 

\end{figure}

\paragraph{Statistical modeling strategy}
We estimate linear mixed-effects models for the analyses reported in  this paper. We fit these models with the lme4 package in R \cite{bates2015fitting}. Models include random-effect  for legislators, which account for repeated observations from the same member across Congressional sessions or calendar year-quarters and allow baseline EMI levels to vary across members. We include fixed-effects for time periods, such as Congress or year-quarter to adjust for period-specific shocks. 
We report cluster-robust standard errors with the bias-reduced cluster-robust variance estimator (CR2) adjustment, with clustering at the legislator level. This adjustment allows for inference that is less dependent on assumptions about the within-legislator covariance structure while correcting for finite-sample bias associated with cluster leverage. \cite{bell2002bias,pustejovsky2018small}.
We use this modeling strategy across the analyses  that follow.

\paragraph{Statistical analyses of the relationship between EMI and ideology}

We examine the relationship between EMI and ideological extremity while controlling for other characteristics of members of Congress. We measure ideological extremity as the absolute value of the first dimension of the DW-NOMINATE (Dynamic Weighted NOMINAl Three-step Estimation) score \cite{poole2011ideology, lewis2026voteview}. DW-NOMINATE estimates ideological positions from roll-call voting behaviour and places members of Congress on a common spatial scale. The first dimension captures the primary liberal-conservative dimension in congressional voting. We use the absolute value to measure distance from the ideological center, with larger values indicating more extreme ideological positions.

We specify the mixed-effects models as follows:

\begin{equation*}
\text{EMI}_{i,t} = \beta_0 + \beta_1 \text{ideology}_{i,t} + \boldsymbol{\beta}\mathbf{X}_{i,t} + \boldsymbol{\gamma}\mathbf{F}_{t} + u_i + \varepsilon_{i,t},
\end{equation*}

\noindent where $i$ indexes members of Congress and $t$ indexes congressions. $\text{ideology}_{i,t}$ denotes the DW-NOMINATE dimension one score, $\mathbf{X}_{i,t}$ denotes the covariates included in a given specification, and $\mathbf{F}_{t}$ denotes the fixed effects included in that specification. The term $u_i$ denotes a legislator random intercept, and $\varepsilon_{i,t}$ denotes the residual error term.

We estimate three nested model specifications. The first specification includes party affiliation, chamber, and Congress fixed effects. Party affiliation accounts for likely differences in partisan communication style. Chamber indicators account for institutional differences between the House and Senate.
Congress fixed effects absorb period-specific shocks.

The second specification adds gender, opposition status, and topic composition. Gender accounts for demographic differences that may explain differences in communication style. We include two opposition indicators: whether a member belongs to the minority party in the legislative chamber and whether a member belongs to the party opposing the president. These variables account for differences in institutional and strategic position. Members in chamber or presidential opposition may face different incentives to criticize, justify, or frame policy claims than members aligned with the chamber majority or the presidency. Topic composition accounts for differences in the substantive content of speeches, since some policy domains may invite more evidence-oriented language than others. We measure topic composition as the proportion of speech content assigned to Comparative Agendas Project (CAP) policy categories using a classifier introduced in prior work \cite{Aroyehun2025Computational}. We omit the residual ``Other'' category as the reference category to avoid perfect collinearity across topic proportions.

The third model specification builds on the second and adds covariates available from 1973 to 2024. Seniority and first-term status account for legislative experience and adaptation to congressional speech norms. Committee chair, subcommittee chair, majority-party leadership, and minority-party leadership account for institutional responsibilities and agenda-setting roles, which may influence the nature of speeches. Vote share in the previous election accounts for electoral security, since members in safer seats may face different communication incentives than members in competitive districts. Prior service in a state legislature accounts for earlier legislative experience that may also shape congressional speeches. We derive these covariates from the voteview \cite{lewis2026voteview} and LES \cite{volden2014legislative, volden2018legislative} datasets.

Table \ref{tab:emi_xteristics} reports the results across the three specifications. EMI remains negatively associated with ideological extremity across models, indicating that more ideologically extreme legislators use less evidence-oriented language even after controlling for partisan, institutional, temporal, topical, demographic, prior legislative experience, leadership, and electoral factors.

\begin{table}[htbp]
    \centering
    \caption{Regression results for correlates of EMI. Fits start in 1873 and ends in 2024 (43rd to 118th Congress) for Models 1 and 2. Model 3 starts in 1973 (93rd Congress) using available data to account for more comprehensive characteristics of congressional members. Coefficients in bold are significant at the 0.05 level. Standard errors are clustered at the legislator level. 95\% confidence intervals in square brackets. 
    }
    \label{tab:emi_xteristics}
    \begin{tabular}{cccc}
\hline
& EMI & EMI & EMI \\
& (1) & (2) & (3) \\ \hline
Ideology & \textbf{-0.116} & \textbf{-0.072} & \textbf{-0.148} \\
& [-0.133, -0.098] & [-0.086, -0.057] & [-0.175, -0.121] \\
& p=1.117e-38 & p=1.570e-21 & p=1.151e-26 \\
Republican & \textbf{0.063} & 0.016 & \textbf{-0.062} \\
& [0.029, 0.096] & [-0.012, 0.044] & [-0.111, -0.012] \\
& p=2.443e-04 & p=0.269 & p=0.014 \\
Senate & \textbf{-0.065} & \textbf{-0.088} & \textbf{-0.063} \\
& [-0.101, -0.029] & [-0.120, -0.055] & [-0.120, -0.006] \\
& p=4.181e-04 & p=1.013e-07 & p=0.030 \\
Chamber opposition &  & \textbf{-0.139} & \textbf{-0.105} \\
&  & [-0.155, -0.123] & [-0.131, -0.080] \\
&  & p=5.295e-63 & p=5.072e-16 \\
Presidential opposition &  & \textbf{-0.016} & \textbf{-0.080} \\
&  & [-0.032, -0.001] & [-0.103, -0.057] \\
&  & p=0.036 & p=1.370e-11 \\
Female &  & 0.009 & 0.058 \\
&  & [-0.048, 0.065] & [-0.014, 0.130] \\
&  & p=0.768 & p=0.116 \\
Seniority (terms served) &  &  & 0.005 \\
&  &  & [-0.020, 0.030] \\
&  &  & p=0.698 \\
First term &  &  & 0.023 \\
&  &  & [-0.013, 0.059] \\
&  &  & p=0.208 \\
Committee chair &  &  & \textbf{0.371} \\
&  &  & [0.311, 0.431] \\
&  &  & p=1.520e-33 \\
Subcommittee chair &  &  & \textbf{0.240} \\
&  &  & [0.205, 0.274] \\
&  &  & p=6.718e-42 \\
Majority party leadership &  &  & -0.017 \\
&  &  & [-0.097, 0.064] \\
&  &  & p=0.685 \\
Minority party leadership &  &  & \textbf{0.133} \\
&  &  & [0.047, 0.218] \\
&  &  & p=0.002 \\
Vote share (last election) &  &  & \textbf{0.016} \\
&  &  & [0.001, 0.031] \\
&  &  & p=0.042 \\
Served in state leg. &  &  & -0.007 \\
&  &  & [-0.053, 0.040] \\
&  &  & p=0.781 \\
Topic FE & No & Yes & Yes \\
Congress FE & Yes & Yes & Yes \\
Num.Obs. & 37686 & 37686 & 13682 \\
R2 Marg. & 0.156 & 0.339 & 0.384 \\
R2 Cond. & 0.526 & 0.587 & 0.625 \\
\hline
\end{tabular}

\end{table}

\paragraph{Statistical analyses of EMI across Twitter and congressional floor speeches}
We estimate three sets of mixed-effects models to examine whether EMI generalizes across communication settings and whether the communication platform itself predicts EMI. All three analyses use legislator-quarter observations from 2013 to 2022 and restrict the sample to members of Congress with non-missing EMI scores for both Twitter posts and congressional floor speeches in the same quarter.

First, we test whether EMI on Twitter predicts EMI in floor speeches. The outcome is floor-speech EMI for member $i$ in quarter $t$, and the main predictor is Twitter EMI for the same member and quarter. Second, we estimate the reverse specification, using Twitter EMI as the outcome and floor-speech EMI as the main predictor. These two models assess whether members who use more evidence-oriented language in one setting also tend to do so in the other.

We specify these cross-platform models as:

\begin{equation*}
\text{EMI}^{A}_{i,t} =
\beta_0 +
\beta_1 \text{EMI}^{B}_{i,t} +
\boldsymbol{\beta}\mathbf{X}_{i,t} +
\boldsymbol{\gamma}\mathbf{F}_{t} +
u_i +
\varepsilon_{i,t},
\end{equation*}
where $i$ indexes members of Congress and $t$ indexes calendar quarters. $\text{EMI}^{A}_{i,t}$ denotes EMI in the outcome platform, and $\text{EMI}^{B}_{i,t}$ denotes EMI in the other platform. $\mathbf{X}_{i,t}$ denotes covariates included in a given specification, $\mathbf{F}_{t}$ denotes fixed effects, $u_i$ denotes a legislator random intercept, and $\varepsilon_{i,t}$ denotes the residual error term.

Third, we estimate a model to test whether EMI differs systematically between Twitter and congressional floor speeches. We restructure the data to the legislator-quarter-platform level and model EMI as a function of an indicator for Twitter posts, with floor speeches as the reference category. This model estimates whether the same members use more or less evidence-oriented language on Twitter than on the congressional floor.

We specify the model as:

\begin{equation*}
\text{EMI}_{i,t,p} =
\beta_0 +
\beta_1 \text{Twitter}_{p} +
\boldsymbol{\beta}\mathbf{X}_{i,t} +
\boldsymbol{\gamma}\mathbf{F}_{t} +
u_i +
\varepsilon_{i,t,p},
\end{equation*}
where $p$ indexes the communication platform. $\text{Twitter}_{p}$ equals one for Twitter and zero for congressional floor. 

Across these models, we include covariates that account for party affiliation, chamber, chamber opposition, and presidential opposition. These controls account for partisan, institutional, and strategic differences that may shape communication across platforms. We include year-quarter fixed effects to absorb common shocks and platform-wide changes over time, such as election cycles, major political events, or changes in the communication environment. Tables \ref{tab:twitter_predict_congress}, \ref{tab:congress_predict_twitter}, and \ref{tab:medium_effect_emi} report the results.

\begin{table}[htbp]
    \centering
    \caption{EMI on Twitter is a significant predictor of EMI on Congressional floor. Models span 2013 Q1 to 2022 Q4. Coefficients in bold are significant at the 0.05 level. Standard errors are clustered at the legislator level. 95\% confidence intervals in square brackets.}
    \label{tab:twitter_predict_congress}
    \begin{tabular}{lll}
\hline
& EMI (floor) & EMI (floor) \\
& (1) & (2) \\ \hline
EMI (Twitter) & \textbf{0.362} & \textbf{0.288} \\
& [0.313, 0.411] & [0.235, 0.341] \\
& p=1.351e-46 & p=2.668e-26 \\
Chamber opposition &  & \textbf{-0.281} \\
&  & [-0.359, -0.204] \\
&  & p=1.605e-12 \\
Presidential opposition &  & \textbf{-0.091} \\
&  & [-0.153, -0.030] \\
&  & p=0.003 \\
Senate &  & \textbf{-0.330} \\
&  & [-0.490, -0.169] \\
&  & p=5.725e-05 \\
Republican &  & \textbf{-0.279} \\
&  & [-0.406, -0.151] \\
&  & p=1.839e-05 \\
Year-Quarter FE & Yes & Yes \\
Num.Obs. & 4191 & 4191 \\
R2 Marg. & 0.174 & 0.203 \\
R2 Cond. & 0.560 & 0.575 \\
\hline
\end{tabular}

\end{table}

\begin{table}[htbp]
    \centering
    \caption{EMI on Congressional floor is a significant predictor of EMI on Twitter. Models span 2013 Q1 to 2022 Q4. Coefficients in bold are significant at the 0.05 level. Standard errors are clustered at the legislator level. 95\% confidence intervals in square brackets.}
    \label{tab:congress_predict_twitter}
    \begin{tabular}{lll}
\hline
& EMI (Twitter) & EMI (Twitter) \\
& (1) & (2) \\ \hline
EMI (floor) & \textbf{0.357} & \textbf{0.233} \\
& [0.309, 0.405] & [0.189, 0.276] \\
& p=8.539e-47 & p=3.189e-25 \\
Chamber opposition &  & \textbf{-0.369} \\
&  & [-0.466, -0.273] \\
&  & p=9.425e-14 \\
Presidential opposition &  & \textbf{-0.540} \\
&  & [-0.613, -0.467] \\
&  & p=1.114e-46 \\
Senate &  & 0.017 \\
&  & [-0.228, 0.262] \\
&  & p=0.893 \\
Republican &  & -0.054 \\
&  & [-0.183, 0.075] \\
&  & p=0.415 \\
Year-Quarter FE & Yes & Yes \\
Num.Obs. & 4191 & 4191 \\
R2 Marg. & 0.168 & 0.243 \\
R2 Cond. & 0.574 & 0.656 \\
\hline
\end{tabular}

\end{table}

\begin{table}[htbp]
    \centering
    \caption{Association between channel of communication and EMI across Twitter and Congressional floor. Models span 2013 Q1 to 2022 Q4. Coefficients in bold are significant at the 0.05 level. Standard errors are clustered at the legislator level. 95\% confidence intervals in square brackets.}
    \label{tab:medium_effect_emi}
    \begin{tabular}{lll}
\hline
& EMI & EMI \\
& (1) & (2) \\ \hline
Twitter & \textbf{-0.150} & \textbf{-0.150} \\
& [-0.242, -0.057] & [-0.242, -0.057] \\
& p=0.002 & p=0.002 \\
Legislative opposition &  & \textbf{-0.434} \\
&  & [-0.510, -0.357] \\
&  & p=1.652e-28 \\
Presidential opposition &  & \textbf{-0.416} \\
&  & [-0.468, -0.363] \\
&  & p=1.115e-53 \\
Senate &  & -0.246 \\
&  & [-0.526, 0.034] \\
&  & p=0.085 \\
Republican &  & \textbf{-0.229} \\
&  & [-0.351, -0.107] \\
&  & p=2.295e-04 \\
Year-Quarter FE & Yes & Yes \\
Num.Obs. & 8382 & 8382 \\
R2 Marg. & 0.032 & 0.137 \\
R2 Cond. & 0.438 & 0.495 \\

\hline
\end{tabular}

\end{table}

\paragraph{Statistical analyses of the relationship between EMI and legislative effectiveness}
We examine whether EMI predicts legislative effectiveness. The outcome is the LES score, a member-Congress measure of how successful they are in advancing legislation through the lawmaking process \cite{volden2014legislative, volden2018legislative}. 
The LES is a widely used metric designed to assess the success of individual members of Congress at advancing legislation through the legislative process during a two-year Congressional session. It accounts for significant actions such as bill introductions, committee referrals, and floor votes, while weighing their importance in the legislative process. The LES reflects both the quantity and quality of efforts of individual legislators, as well as their influence and effectiveness within the legislative process.
The main predictor is floor-speech EMI, averaged for each member in each two-year Congress. The analysis covers the 93rd to the 118th Congress (1973 to 2024) corresponding to the period covered by the LES dataset.

We specify the LES models as follows:

\begin{equation*}
\text{LES}_{i,t} =
\beta_0 +
\beta_1 \text{EMI}_{i,t} +
\boldsymbol{\beta}\mathbf{X}_{i,t} +
\boldsymbol{\gamma}\mathbf{F}_{t} +
u_i +
\varepsilon_{i,t},
\end{equation*}
where $i$ indexes members of Congress and $t$ indexes Congresses. $\text{LES}_{i,t}$ denotes legislative effectiveness for each member in a given Congress, $\text{EMI}_{i,t}$ denotes the average EMI score derived from speeches for a member in a given Congress, $\mathbf{X}_{i,t}$ denotes covariates included in each specification, and $\mathbf{F}_{t}$ denotes fixed effects included in each specification. The term $u_i$ denotes a legislator random intercept, and $\varepsilon_{i,t}$ denotes the residual error term.
We estimate three specifications guided by prior work on legislative effectiveness.

We estimate three LES models with the exact covariate sets reported in Table \ref{tab:emi_les}. The baseline specification includes established predictors from the LES literature, including ideology, party position, chamber, opposition status, gender, race and ethnicity, seniority, first-term status, committee and subcommittee chair positions, party leadership roles, vote share in the previous election, and prior service in a state legislature. These variables capture factors thought to be associated with legislative effectiveness in prior work.

The second specification adds each member average EMI score to test whether evidence-oriented communication predicts legislative effectiveness beyond the established predictors. A third specification further adds topic composition and communication frequency and volume, measured using speech count and token count. Topic composition accounts for differences in the policy domains discussed by members, while communication volume accounts for differences in patterns number of and extent of speech making on the congressional floor.
Table \ref{tab:emi_les} reports the results across the three specifications.
\begin{table}[htbp]
    \footnotesize
    \centering
    \caption{Relationship between EMI and LES (Legislative Effectiveness score). Fits start in 1973 and end in 2024 (93rd to 118th Congress). Coefficients in bold are significant at the 0.05 level. Standard errors are clustered at the legislator level. 95\% confidence intervals in square brackets.}
\label{tab:emi_les}

    \begin{tabular}{llll}
\hline
& LES & LES & LES \\
& (1) & (2) & (3) \\ \hline
EMI &  & \textbf{0.163} & \textbf{0.129} \\
&  & [0.142, 0.185] & [0.106, 0.152] \\
&  & p=5.820e-51 & p=2.696e-28 \\
Ideology & \textbf{-0.037} & 0.003 & \textbf{-0.047} \\
& [-0.066, -0.007] & [-0.027, 0.032] & [-0.074, -0.020] \\
& p=0.015 & p=0.857 & p=6.255e-04 \\
Chamber opposition & \textbf{0.049} & \textbf{0.067} & 0.036 \\
& [0.011, 0.087] & [0.029, 0.104] & [-0.001, 0.073] \\
& p=0.011 & p=4.827e-04 & p=0.060 \\
Presidential opposition & -0.002 & 0.013 & 0.004 \\
& [-0.036, 0.032] & [-0.021, 0.047] & [-0.029, 0.037] \\
& p=0.898 & p=0.465 & p=0.813 \\
Republican & -0.033 & -0.023 & -0.031 \\
& [-0.094, 0.029] & [-0.084, 0.038] & [-0.086, 0.025] \\
& p=0.301 & p=0.457 & p=0.279 \\
Female & \textbf{0.151} & \textbf{0.141} & 0.049 \\
& [0.066, 0.235] & [0.059, 0.224] & [-0.029, 0.128] \\
& p=4.599e-04 & p=7.731e-04 & p=0.218 \\
Senate & \textbf{-0.099} & \textbf{-0.098} & \textbf{-0.477} \\
& [-0.175, -0.022] & [-0.174, -0.022] & [-0.558, -0.396] \\
& p=0.012 & p=0.011 & p=1.812e-30 \\
Latino(a) & 0.028 & 0.025 & -0.001 \\
& [-0.136, 0.191] & [-0.133, 0.183] & [-0.142, 0.139] \\
& p=0.738 & p=0.757 & p=0.985 \\
African American & \textbf{-0.162} & \textbf{-0.138} & \textbf{-0.099} \\
& [-0.282, -0.041] & [-0.258, -0.018] & [-0.210, 0.013] \\
& p=0.009 & p=0.024 & p=0.083 \\
Seniority (terms served) & \textbf{-0.132} & \textbf{-0.134} & \textbf{-0.168} \\
& [-0.162, -0.102] & [-0.163, -0.105] & [-0.197, -0.139] \\
& p=2.238e-18 & p=1.180e-19 & p=1.490e-30 \\
First term & \textbf{-0.193} & \textbf{-0.190} & \textbf{-0.157} \\
& [-0.248, -0.138] & [-0.245, -0.136] & [-0.210, -0.103] \\
& p=6.887e-12 & p=9.963e-12 & p=7.929e-09 \\
Committee chair & \textbf{0.111} & 0.046 & -0.014 \\
& [0.034, 0.188] & [-0.030, 0.122] & [-0.088, 0.060] \\
& p=0.005 & p=0.236 & p=0.710 \\
Subcommittee chair & 0.033 & -0.016 & -0.038 \\
& [-0.016, 0.083] & [-0.065, 0.032] & [-0.084, 0.008] \\
& p=0.183 & p=0.508 & p=0.105 \\
Majority party leadership & \textbf{0.216} & \textbf{0.234} & \textbf{0.239} \\
& [0.092, 0.339] & [0.115, 0.354] & [0.126, 0.352] \\
& p=6.285e-04 & p=1.208e-04 & p=3.386e-05 \\
Minority party leadership & 0.017 & -0.001 & -0.036 \\
& [-0.103, 0.137] & [-0.116, 0.114] & [-0.149, 0.077] \\
& p=0.783 & p=0.990 & p=0.529 \\
Vote share (last election) & \textbf{-0.028} & \textbf{-0.028} & \textbf{-0.024} \\
& [-0.048, -0.008] & [-0.048, -0.009] & [-0.043, -0.005] \\
& p=0.006 & p=0.005 & p=0.012 \\
Served in state leg & 0.026 & 0.026 & 0.022 \\
& [-0.030, 0.081] & [-0.029, 0.080] & [-0.027, 0.072] \\
& p=0.365 & p=0.355 & p=0.383 \\
Speech count (log) &  &  & 0.177 \\
&  &  & [0.116, 0.239] \\
&  &  & p=1.683e-08 \\
Token count (log) &  &  & 0.128 \\
&  &  & [0.070, 0.186] \\
&  &  & p=1.553e-05 \\
Topic FE & No & No & Yes \\
Congress FE & Yes & Yes & Yes \\
Num.Obs. & 13682 & 13682 & 13682 \\
R2 Marg. & 0.024 & 0.045 & 0.121 \\
R2 Cond. & 0.314 & 0.326 & 0.334 \\

\hline
\end{tabular}

\end{table}

\backmatter

\clearpage
\section*{Acknowledgments}

SL acknowledges financial support from 
the European Research Council
(ERC Advanced Grant 101020961 PRODEMINFO).
DG is also a beneficiary of the ERC Advanced Grant 101020961 PRODEMINFO. SA is supported by PRODEMINFO.

\subsection*{Author Contributions Statement} SA, SL, and DG conceptualised the research. SA collected the data. SA developed the text analysis pipeline. SA performed the statistical analyses. SA prepared the initial draft of the manuscript. All authors contributed to preparing and editing the final version of the manuscript. 

\section*{Competing Interests Statement} Authors have no competing interests.

\clearpage
\bibliography{refs}

\end{document}